\documentclass[runningheads]{llncs}
\usepackage[T1]{fontenc}
\usepackage{graphicx}
\usepackage{array}
\begin{document}
\title{Demand Forecasting using Long Short-Term Memory Neural Networks\thanks{Supported by inovex GmbH.}}
\author{Marta Go\l{}\k{a}bek\inst{1,2} \and
Robin Senge\inst{2} \and
Rainer Neumann\inst{1}}
% \orcidID{0000-0002-8696-7639}

%
\authorrunning{M. Go\l{}\k{a}bek et al.}
\renewcommand{\and}{\\} % ewentualnie do usuniecia
\institute{Karlsruhe University of Applied Sciences, 76133 Karlsruhe, Germany \and
inovex GmbH, 76131 Karlsruhe, Germany \\
\email{mgolabek94@gmail.com}\quad\email{rsenge@inovex.de}\quad\email{rainer.neumann@hs-karlsruhe.de}}
%\url{http://www.springer.com/gp/computer-science/lncs}} %\and
\maketitle              % typeset the header of the contribution
\begin{abstract}
In this paper we investigate to what extent long short-term memory neural networks (LSTMs) are suitable for demand forecasting in the e-grocery retail sector. 
For this purpose, univariate as well as multivariate LSTM-based models were developed and tested for 100 fast-moving consumer goods in the context of a master's thesis.

On average, the developed models showed better results for food products than the comparative models from both statistical and machine learning families. Solely in the area of beverages random forest and linear regression achieved slightly better results. This outcome suggests that LSTMs can be used for demand forecasting at product level.

The performance of the models presented here goes beyond the current state of research, as can be seen from the evaluations based on a data set that unfortunately has not been publicly available to date.

\end{abstract}
\section{Introduction}
The aim of this article is to evaluate the suitability of long short-term memory neural networks (LSTMs) for demand forecasting in the field of electronic food retailing (e-grocery). The focus lies on fast-moving consumer goods, i.e. products that are traded in short cycles with only short storage time and at typically low prices. The models should be able to make a forecast for an entire working week, i.e. a multi-step forecast based on days.

For this purpose, both univariate and multivariate models are developed and evaluated in the following. While the univariate models only use past demand data, the multivariate models additionally consider exogenous characteristics such as weekdays or prices. 

In addition, it is investigated how the quality of prediction can be improved by suitable pre-training and how substitutive and complementary dependencies between products (substitutes and complementary goods) can be used for further improvement in the form of parallel prediction.

\subsection{Motivation}
The e-grocery retail is a challenging business. The reason behind this is twofold, the value of goods being sold is low, whereas the additional costs caused by online retailing are high \cite{fernie2018}. Therefore, it is crucial to identify avoidable costs along the supply chain and reduce them. A great share of such costs arises either from excessive or insufficient coverage. Since demand forecasting plays a vital role in the control of retailers' supply chains, an improvement of its accuracy can contribute to significant cost reduction \cite{fernie2018,Bandara2019,hasin2011ann,patak2015demand}.

LSTMs have a few promising properties with regard to the product-level demand forecasting. First of all, they are naturally suited for modelling time series. This is due to their ability to capture dependency in a sequential context and preserve past information as they progress through the subsequent time steps in a series \cite{Bandara2018}. Furthermore, LSTMs accept multivariate inputs. Thus, not only historical demand data but also further determinants of demand such as price or promotion can be considered at the time of predicting \cite{Bandara2019}. Finally, LSTMs are capable of performing predictions for multiple products in parallel. In this way, complementary and substitutive effects between the goods can be learned by the model.

Despite this architectural suitability, the forecasting community remains hesitant about the application of LSTMs to time series forecasting. This is mostly owing to a lack of empirical evidence as well as an absence of evaluation metrics and standard benchmarks \cite{Bandara2018}.

\subsection{Business Setting}
The data set used for the research was provided by one of Germany's largest grocery retailers on account of its wide interest in accurate demand forecasting for its e-grocery solution.

Order fulfilment takes place in a distribution warehouse, so it is expected to have the ordered products in stock at time of request. The distribution warehouse is supplied by the responsible central warehouse. The product quantity ordering between the warehouses is optimised using an already existing optimisation system. However, in order to perform the optimisation, the system requires a demand forecast.

There is an existing demand forecast solution that facilitates several statistical and machine learning (ML) models. Nonetheless, in order to ensure high accuracy of the forecasts, the development of new forecasting models is an ongoing process. The existing models function as benchmarks for the developed LSTM-based models.
%Therefore, the retailer is interested in evaluating the suitability of LSTMs for demand forecasting.
\section{Related Research}
The potential of LSTMs for demand forecasting in retail has already been identified by the research community. Thus, there exist plenty of related research papers. In the following, prior work is summarized and a research gap is identified.
% that is addressed in this paper

First of all, many of the demand forecasting models developed with LSTMs are still univariate nowadays, e.g. models built in \cite{Wang2018LSTMSalesForecast} and \cite{goyal2018solution}. This approach does not exploit the substantial advantage of LSTMs over univariate statistical methods to be able to consider exogenous variables. The latter is essential as demand is strongly affected by a great number of factors such as price or promotions \cite{william2012managerial}.

Furthermore, many studies train LSTM on a single time series, e.g. \cite{Wang2018LSTMSalesForecast}, \cite{goyal2018solution} and \cite{Helmini2019}. However, since LSTMs are prone to overfitting when provided with insufficient data, it may lead to poor performance of a model. As shown in \cite{smyl2016forecasting}, training across related time series can solve the problem of scarce data. A further point of criticism is the fact that many of the models created in related work perform solely one-step forecasts, e.g. in \cite{Wang2018LSTMSalesForecast} and \cite{Helmini2019}.
%Notwithstanding, it is of interest how LSTMs perform on longer forecast horizons.

Further, the work presented in \cite{Bandara2019}, which together with \cite{Helmini2019} represents the most advanced approach to LSTM-based demand forecasting found in related research, fails to model yearly seasonality owing to the too short time series being used for training.
%In order to ensure that models being developed here are able to learn yearly seasonalities, they are supplied with two full years of demand data.

All in all, development of a multi-step multivariate model trained across related time series and evaluation of its performance against state-of-the-art forecasting methods makes for a meaningful contribution to demand forecasting in retail.
%Such an empirical evidence is particularly important since due to its current lack, the forecasting community remains hesitant about the application of LSTMs to time series forecasting \cite{Bandara2018}.

Moreover, this work takes up on the improvement suggested in the outlook of the paper \cite{Helmini2019} and provides the model not only with future-oriented information for the day being predicted but also beyond that day (further future features). Among others the model is provided with an information whether the store is going to be open the following day as it strongly affects the customer behaviour on a day in question.

Finally, cross-series information has so far only been used for increasing the data volume, thus leaving the substitutive and complementary effects unaddressed. This work tries to utilise their potential for demand forecasting by developing a model that produces forecasts for substitutes and complementary goods in parallel. Incorporating the listed improvements is likely to enhance the performance of the hitherto presented approaches.
\section{Data Set and Data Preprocessing}
The data set provided for the research consisted of daily sales information of 100 fast-moving consumer goods in five distribution warehouses collected over two years and five months. Nonetheless, days on which the store was closed such as public holidays were not included.

The data set already comprised numerous features as they had been engineered for the existing demand forecasting models. The only feature that had to be added was information about previous demand. So far demand data had only been used as an output. Thus, each record was augmented by feature indicating the quantity demanded on the previous working day. The categorical features such as day of the week were one-hot encoded. There were almost no missing values in the data set. They occurred solely for the feature \emph{price} and were replaced by the mean price.

There seems to be a consensus regarding necessity of normalization of the series prior training using LSTMs. This preprocessing step was applied in all reviewed related works. Also an empirical experiment conducted on the provided data set confirmed the need of normalization. It is particularly important when training across multiple related time series as the ranges of values may differ significantly between the series. Since the trend in the time series was rather weak, the series were normalized in one shot using the min-max scaler \cite{min_max_scaler_scikit}. However, moving window normalization may be a better choice if time series have a stronger trend.

On the contrary, there is no such unanimity with regard to the need of deseasonalization in the context of LSTMs. The ability of LSTMs to approximate any arbitrary function suggests that they are capable of modeling seasonality on their own. This used to be a common opinion among researchers in the past. However, more recent studies argue that conducting deseasonalization prior to training contributes to more accurate forecasts \cite{Bandara2018}. The authors of \cite{Bandara2018} advise to train both a model with and without deseasonalization, and then select the one with a better performance on the validation set. However, they also state that the deseasonalization step may not be needed, when the model is provided with calendar features or/and time series show homogenous seasonality patterns. For instance, deseasonalization is omitted in \cite{Bandara2019} and \cite{Helmini2019}.
%, which represent the most advanced approaches to LSTM-based demand forecasting in retail found in related research.

Since the data set used here included calendar features and the training was conducted either on a single time series or across related time series, which implies homogeneous seasonality patterns, deseasonalization was likely to be unnecessary. Moreover, the models were supplied with two full years of training data, which should additionally enhance their ability to learn seasonalities throughout a year. Therefore, no deseasonalization was performed.

LSTMs require fixed-size input-output pairs for training and the common strategy to achieve that is the moving window (MW) approach. This technique was also applied here. The input window width was determined experimentally, whereas the output window width amounted to the forecast horizon.

Finally, the data set was split. First two years were allocated to the training set. The next three months were used as a validation set based on which hyperparameters were tuned. The most recent two months were used for the assessment of the model predictive performance and comparison with the benchmarks.

\section{Network Architecture}
Many of the reviewed related works such as \cite{Bandara2019} and \cite{Helmini2019} used LSTM with peephole connections. This variant of LSTM is known to be better at learning the timing of a sequence \cite{LSTMvariants2020,gers2002learning}. However, since days on which the store was closed were not included in the data set, counting of the time steps might lead to the deterioration of the network's performance. Therefore, the developed LSTM model was not supposed to count the steps on its own, but rather rely on the calendar features provided as exogenous variables. Consequently, traditional LSTM as described in \cite{LSTMvariants2020} is a better choice here.
% For instance, assuming that demand on Mondays is particularly high, if the model would count the time steps instead of being provided with a feature day of the week and some days are missing, e.g. because there were public holidays and the store was closed, it will disturb the learning process.

Based on the related works, i.a. \cite{Bandara2019} and \cite{goyal2018solution}, a common architectural choice seems to be a combination of a single LSTM layer with a fully connected output layer with a linear activation function. The latter fulfills the function of an adaptor that converts the outputs of the LSTM layer to the desired output size, which is equal to the forecast horizon. In some papers, the LSTM layer was additionally followed by one or more fully connected layers with a nonlinear activation function. Hereinafter referred to as nonlinear layers. It helps to capture the remaining nonlinearities \cite{Helmini2019}. This work takes up on this architecture variant, however the number of hidden layers following the LSTM layer as well as the number of neurons in each of them are hyperparameters tuned at product level.

In order to prevent overfitting, it is recommended to apply some regularization techniques. Therefore, a combination of two well-known regularization methods, dropout and early stopping, is used. Dropout randomly deactivates nodes in the network during training. Hence, a dropout layer is only included in the network architecture during the training phase \cite{Michelucci2018}. Inclusion of a dropout layer after each of hidden layers as well as dropout rate are also hyperparameters. In the paper introducing the dropout layer, it is recommended to apply it to each of the fully connected layers \cite{hinton2012improving}. 

\begin{figure}
\includegraphics[width=\textwidth]{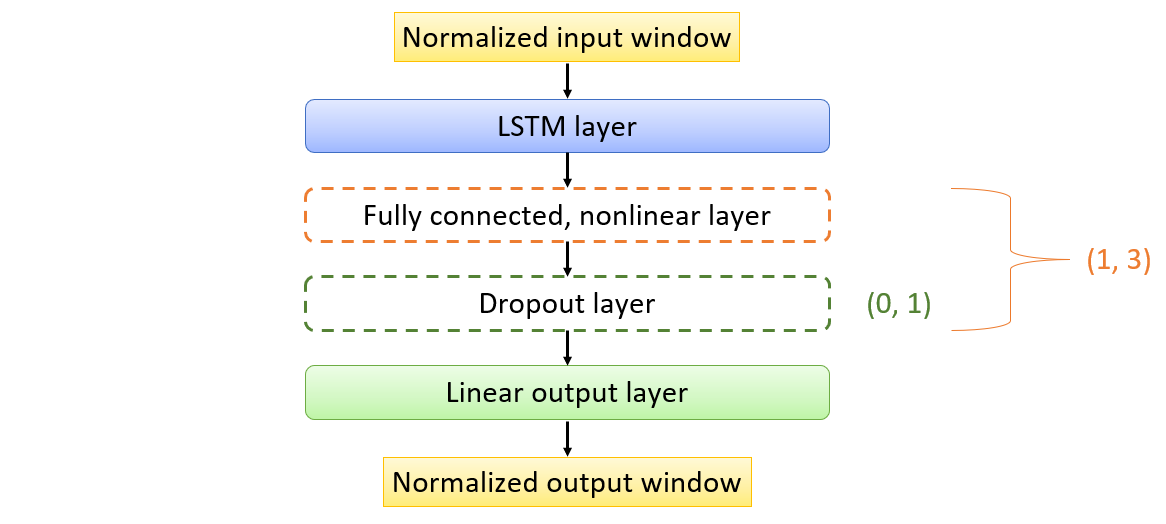}
\caption{Network architecture used during training.} \label{network_architecture_training}
\end{figure}

Since dropout layers are only applied during training, the architecture used in the training phase differs from the one used when performing predictions. The one used during training is schematically illustrated in Figure\,\ref{network_architecture_training}.

Apart from the mentioned hyperparameters regarding the network architecture, there are further hyperparameters that require tuning. All of them are listed in Table\,\ref{hyperparameter_summary}. Some of these hyperparameters are tuned at product level using random search method, whereas the others are determined upfront for the entire assortment based on related work and/or experimentation. Information about the tuning level and method as well as values are also included in Table\,\ref{hyperparameter_summary}.

\begin{table}[ht!]
\caption{Table captions should be placed above the
tables.}\label{hyperparameter_summary}
\begin{tabular}{|>{\centering\arraybackslash}p{5cm}|>{\centering\arraybackslash}p{2cm}|>{\centering\arraybackslash}p{3cm}|>{\centering\arraybackslash}p{3cm}|}
\hline
\textbf{Hyperparameter name} &  \textbf{Level} & \textbf{Method} & \textbf{Values}\\
\hline
Input window width & assortment & related research \& experimentation & 36 time steps\\
\hline
Number of neurons in LSTM & product & random search & (10, 100), step=10\\
\hline
Number of fully connected, nonlinear layers following LSTM & product & random search & (1, 3), step=1\\
\hline
Number of neurons in each of fully connected, nonlinear layers & product & random search & (10, 100), step=10 \\
\hline
Activation function used in each hidden layer & assortment & related research & ReLU\\
\hline
Inclusion of a dropout layer after each of hidden, non-LSTM layers & product & random search & (0, 1), step=1\\
\hline
Dropout rate & product & random search & (0.1, 0.9), step=0.1\\
\hline
Optimizer & assortment & related research & Adam\\
\hline
Learning rate & product & random search & (1e-2, 1e-4), step=1e-1\\
\hline
Loss function & assortment & related research & MSE\\
\hline
Batch size & assortment & related research \& experimentation & 32\\
\hline
Max epochs for early stopping & assortment & related research \& experimentation & 70\\
\hline
Patience for early stopping & assortment & related research \& experimentation & 5\\
\hline
\end{tabular}
\end{table}

\section{Evaluation Framework}
As previously mentioned, the retailer already has a demand forecast solution that encompasses five different models. These are lasso regression (LR), random forest (RF), exponential smoothing (ETS), median previous quarter (MPQ) and median previous quarter calculated based on the day of the week (MDPQ).
%LR and RF use the same feature space as LSTM-based models.
%In order to enable a comparison between them and the LSTM-based models, error metrics specified below are employed.
%the same period, namely the period represented by the test set.

Error metrics used for evaluation are Mean Absolute Error (MAE) and slightly modified Mean Absolute Percentage Error (mMAPE) as proposed by Banadara et al. in \cite{Bandara2019} and expressed by the equation (\ref{mMAPE}). MAE is simple to understand, calculate and widely used. Moreover, it naturally places more weight on fast-moving consumer goods, which is particularly suitable for the retail domain \cite{ma2016}. However, due to the scale-dependency it is not suitable when evaluation encompasses multiple time series \cite{davydenko2016}. Therefore, mMAPE is employed as an additional scale-independent error metric. Similarly to MAE, it is also simple to understand and calculate.

\begin{equation}\label{mMAPE}
mMAPE = \frac{1}{m} \sum_{t=1}^{m} \bigg( \frac{|F_t - A_t|}{1 + |A_t|} \bigg)
\end{equation}

% stands for a single time step out of the forecast horizon
Initially, MAE and mMAPE are calculated for each product with differentiation between lookaheads. The latter stands for a distance between a time step to be forecast and the current time step. Since the forecast horizon amounts to one working week, there are six lookaheads in total. For the purpose of easing the evaluation process, errors are grouped without differentiation between lookaheads and then averaged over all products. This results in two error measures referred to as \emph{overall mean MAE} and \emph{mean mMAPE}. The combination of both scale-dependent and scale-independent metrics provides additional insights into the predictive performance of the models.
% by a model variant

\section{Experimental Design}
At first, a univariate model based solely on past demand data is created. It constitutes a baseline for more advanced models. Then, a multivariate model with only two features, past demand and orders known at the time of prediction, is developed. The latter feature is known for its great predictive potential.

In the next step, further exogenous features are incorporated into the model and the optimal feature space is determined. Besides, the impact of further future features on the model predictive performance is investigated.
% In addition, it was tested whether an LSTM-based model is capable of learning seasonalities when provided with calendar features.

Finally, the approach of pretraining a model using time series belonging to the same product but coming from different distribution warehouses and the approach of parallel forecasting of substitutive and complementary goods are tested. Both of the approaches are expected to contribute to a higher forecast accuracy.

The LSTM-based models are developed in Python 3.6 using Keras, a high-level neural network library, running on top of Tensorflow 2.
%Hyperparameter search is conducted using a tuner developed specifically for Keras and known as Keras Tuner.

\section{Results and Discussion}
\begin{figure}
\includegraphics[width=\textwidth]{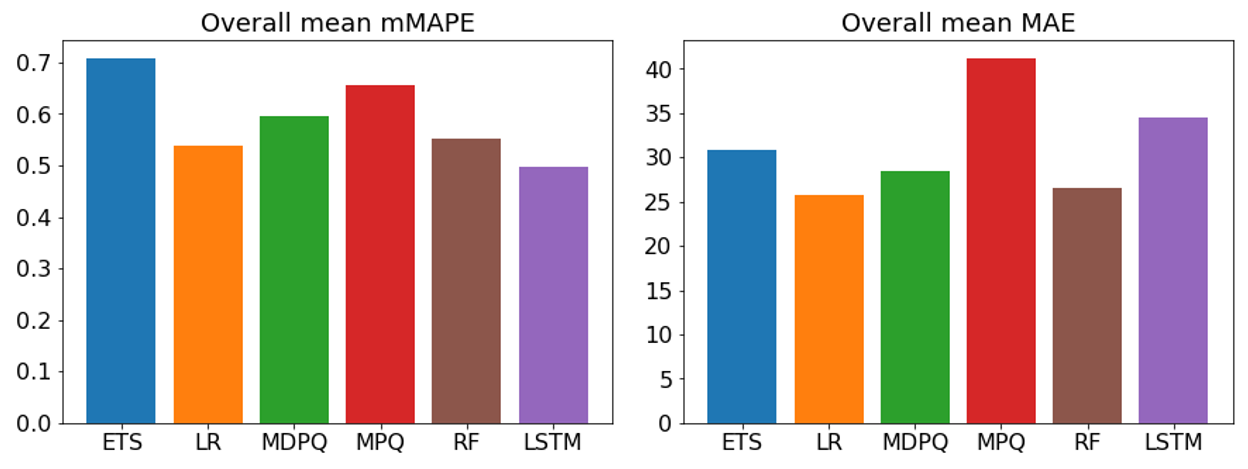}
\caption{Univariate LSTM-based model: comparison with the benchmarks.} \label{univariate_model_results}
\end{figure}

According to the evaluation based on the overall mean mMAPE, the univariate LSTM-based model outperformed all of the benchmark models. However, when the overall mean MAE is considered, the univariate model achieved the second worst result (see Figure\,\ref{univariate_model_results}).

This discrepancy in the model predictive performance when measured using different error metrics may be caused by the fact that the univariate LSTM-based model performs well only for a certain group of products. The good performance indicated by the percentage error metric suggests that this group of products is large. However, the considerably worse results for the absolute error measure show that the products for which the LSTM-based model produces rather poor forecasts are most likely sold in high volumes.
%that there are also products, for which the LSTM-based model causes high errors. Thus, these products are likely to be sold in high volumes, which leads to high absolute errors.

In order to validate this assumption, box plots based on the lookahead-specific mMAPE were created. As expected, for most of the products forecast errors lay between the whiskers of the box plots. Notwithstanding, there were a few extreme outliers, which deteriorated the overall performance of the model. The majority of them appeared to be beverages.

\begin{figure}
\includegraphics[width=\textwidth]{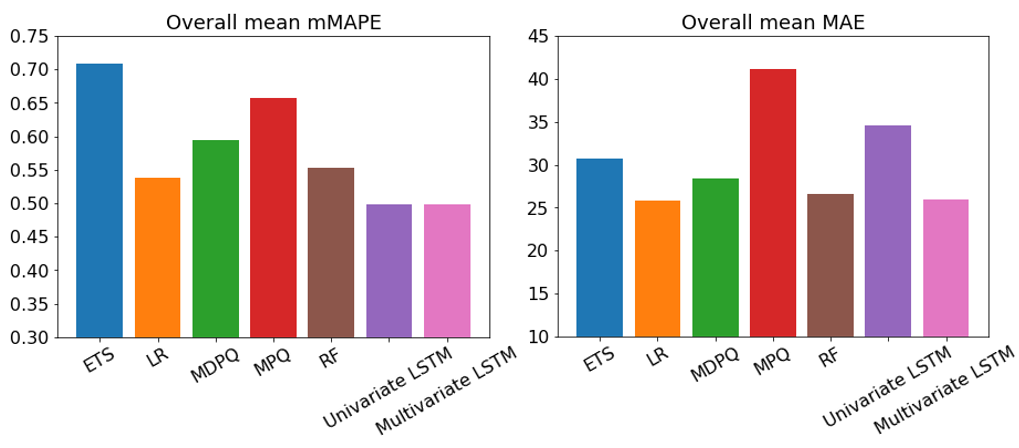}
\caption{Multivariate LSTM-based model with single exogenous feature: comparison with the univariate model and the benchmarks.} \label{multivariate_1_model_results}
\end{figure}

In the next step, the model was augmented by the feature indicating the orders known at the time of prediction. According to the evaluation involving the overall mean mMAPE and mean MAE, this multivariate LSTM-based model outperformed all of the benchmarks (see Figure\,\ref{multivariate_1_model_results}). Based on the percentage error, it achieved similar performance as the univariate model, but the evaluation based on the absolute error indicated a significant improvement over the univariate model. Then it was investigated whether this improvement was caused by a better performance for beverages. Nonetheless, a box plot analysis revealed that beverages are still an issue.
%Therefore, the lower overall mean MAE must be caused by an improvement in the predictive performance for other products.
% Thus, this experiment positively validated the implementation of the multivariate model and confirmed the importance of incorporation of further features.

\begin{figure}
\includegraphics[width=\textwidth]{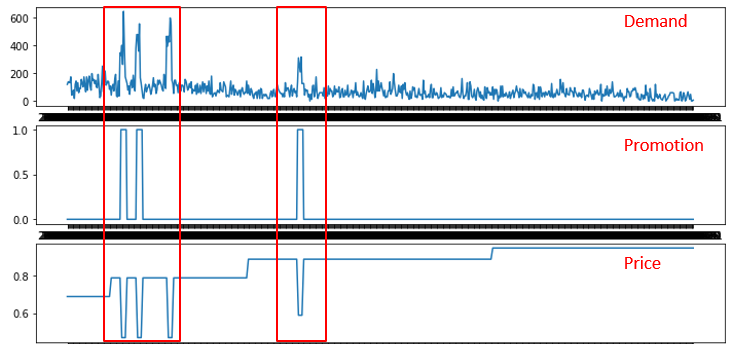}
\caption{Visualization of demand, price and promotion series for an arbitrarily selected outlier.} \label{outlier_ts_visualization}
\end{figure}

Since demand for beverages is characterized by high price elasticity \cite{andreyeva2010impact}, an augmentation of the feature space by features \emph{price} and \emph{promotion} was tested. However, it resulted in deterioration of the model's performance. This outcome was most probably caused by the fact that the available time series tended to include too few examples of considerable price changes for the model to learn them effectively (see Figure\,\ref{outlier_ts_visualization}). Supplying the model with longer time series and thus, more occurrences of price changes may alleviate this problem.
%Besides, missing values for the feature price were replaced by a mean price, which might also contributed to the deterioration in the predictive performance of the model.

Due to the difference in the model predictive performance for food products and beverages, the assortment was divided into these two categories. Then, the optimal feature space was determined separately for each category.

It turned out that the same combination of features achieved the best results for both categories. It included: previous demand, known orders, day of the week, whether tomorrow/the day after tomorrow the store is open and whether tomorrow/the day after tomorrow is a public holiday. The last four features are further future features and they proved to have a positive impact on the model predictive performance.
%Moreover, supplying a model with a single calendar feature, day of the week, turned out to be sufficient for the model to learn the occurring seasonalities on its own.

\begin{figure}[ht!]
\includegraphics[width=\textwidth]{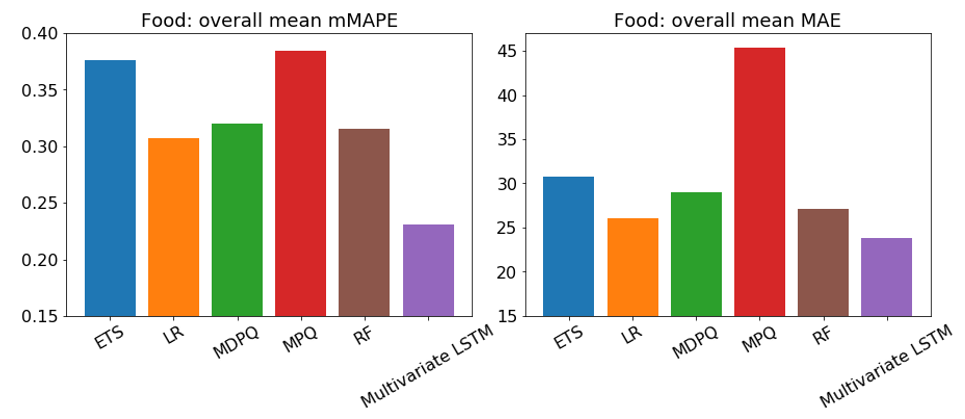}
\caption{Multivariate model with the optimal feature space for the category food: comparison with the benchmark models.} \label{food_multivariate}
\end{figure}

The results of these multivariate models were compared with those of the benchmarks. For the products of the category food, the multivariate LSTM-based model with the optimal feature space outperformed all five benchmark models in the evaluation based on both overall error metrics (see Figure\,\ref{food_multivariate}). At the product level, it achieved the best results for 48 out of 76 products, which constitutes more than 60\% of the items in this category.

\begin{figure}[ht!]
\includegraphics[width=\textwidth]{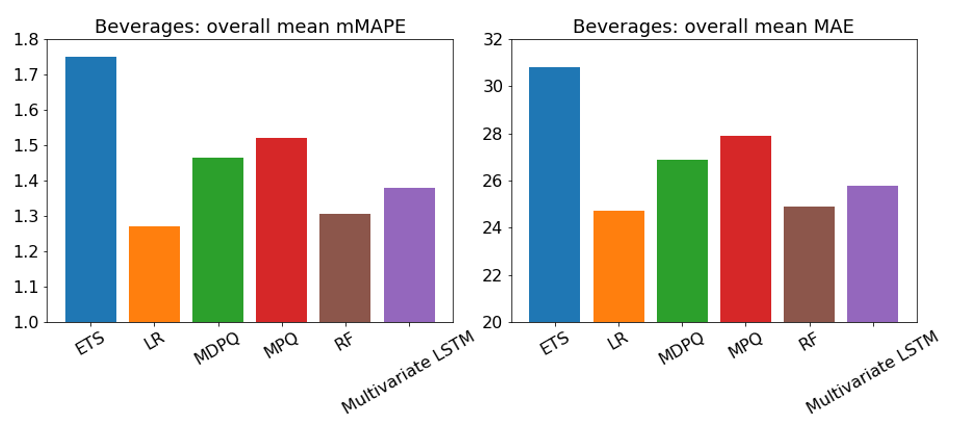}
\caption{Multivariate model with the optimal feature space for the category beverages: comparison with the benchmark models.} \label{beverages_multivariate}
\end{figure}

For the products of the category beverages, the multivariate LSTM-based model with the optimal feature space was slightly outperformed by linear regression and random forest in the evaluation involving overall error metrics (see Figure\,\ref{beverages_multivariate}). Nonetheless, it still performed better than three out of five benchmarks. At the product level, it achieved the best results for 11 out of 24 items, which amounts to about 45\%.

\begin{figure}[ht!]
\includegraphics[width=\textwidth]{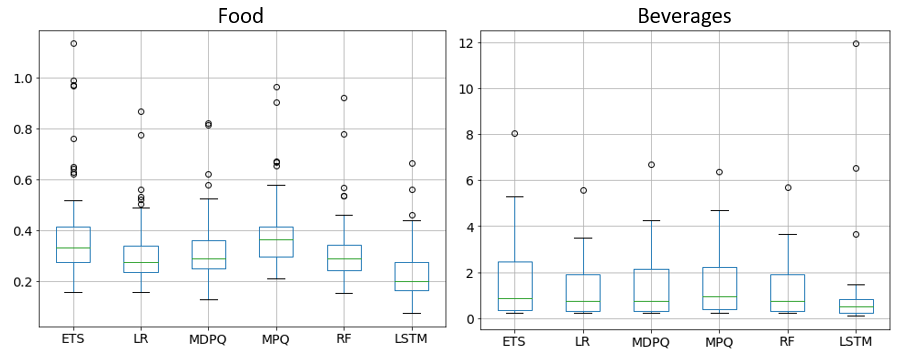}
\caption{Comparison with the benchmark models using box plots based on product-specific mean mMAPE averaged over all lookaheads.} \label{box_plots_final_eval}
\end{figure}

In the next step, the distribution of the errors was investigated using box plots (see Figure\,\ref{box_plots_final_eval}). The diagrams are based on product-specific mean mMAPE averaged over all lookaheads. This analysis showed that the error median obtained by the LSTM-based model is lower than those of the benchmark models for both categories.

The analysis of the final network architectures (after hyperparameter tuning) revealed that the architectures of different products vary significantly in both number of layers and number of neurons. This outcome suggests that tuning an architecture at the product level should be a recommended practice.

Further, the approach of pretraining a model using related time series was tested. Time series used for pretraining belonged to the same product but came from different warehouses. Only time series were used whose spearman's correlation coefficient was at least weak or moderate. Nonetheless, training across related time series did not lead to any improvement in the model's performance. However, this approach may be useful in the setting with shorter time series, more missing values or for slow-moving products.
%Apparently, two and half years of daily sales data are enough for the model to learn to forecast demand for fast-moving consumer goods. However, this approach may be useful in the setting with shorter time series, more missing values or for slow-moving products.

% The experiments were conducted for five different product combinations.
% The parallel forecast may also require a different network architecture.
Finally, the approach of forecasting demand for multiple products in parallel was investigated. On the whole, the parallel forecast achieved worse results than the multivariate LSTM-based model performing forecasts for each product separately. This outcome may be caused by the fact that when predicting multiple products in parallel, the model is being exposed to additional noise. Furthermore, for substitutive and complementary effects to occur price changes are required. The latter tended to be rare in the available time series and not always included in both training and evaluation sets. Therefore, it is recommended to additionally test parallel forecast in an environment with more frequent and more significant price changes. Besides, the products used for experiments were not obvious substitutes or complements. Thus, the parallel forecast may achieve better results when the substitutive and complementary effects are more certain.

\section{Conclusion and Outlook}
In this paper we were able to show that LSTMs are suitable for building a product-level demand forecasting model for electronic food retailing.

The univariate model already showed promising results, but the best results were achieved with the advanced multivariate model. This model provided better forecasts for food products than the 5 benchmark models. For beverages it performed better than 3 of the reference models, but was slightly outperformed by linear regression and random forest approaches.

Training across related time series as well as parallel forecasting might not have unfolded their full potential due to the constraints of the given data set. Therefore, further experiments using a different data set are required. This also holds for products with high price elasticity such as beverages.

Owing to the limited length of the time series provided for the research, the evaluation was conducted only on two months. Consequently, the changing of the seasons could not be considered. Thus, it is required to evaluate the prototype's performance on an entire year. The same holds for hyperparameter tuning. It should also be tested whether extending the input window width as well as employing weather features contributes to more accurate forecasts. But this requires availability of longer time series.

Since accurate forecasts are usually not possible \cite{makridakis2009forecasting}, future work must show, how an estimation of uncertainty can be incorporated into the forecast, e.g. by presenting prediction intervals along with the forecast.

That information might also be useful to minimize the effects of surplus stock and lost profits due to unavailability, since availability is a crucial factor for the shopping process \cite{ulrich2019distributional}.
% They not only enable an assessment of uncertainty associated with each forecast, but also a more through comparison between different forecasting methods \cite{hyndman2018forecasting, chatfield1991prediction}.

Finally, for the practical application it has to be clarified whether the costs of training the highly complex LSTMs are economically viable compared to the savings through improved forecasts.

\end{document}